\documentclass[letterpaper]{article} 
\usepackage{aaai2026}  
\usepackage{times}  
\usepackage{helvet}  
\usepackage{courier}  
\usepackage[hyphens]{url}  
\usepackage{graphicx} 
\usepackage{natbib}  
\usepackage{caption} 
\usepackage{placeins}

\usepackage{xspace}
\usepackage{pifont}
\usepackage{tcolorbox}
\usepackage{tablefootnote}
\usepackage{float}

 \usepackage{booktabs}
\usepackage{multirow}
\usepackage{makecell}

\usepackage[linesnumbered,ruled,vlined]{algorithm2e}
\DontPrintSemicolon
\usepackage{amsmath}
\usepackage{amssymb}

\usepackage{cleveref}

\newcommand{\model}{\textsc{SToLa}\xspace}

\definecolor{myblue}{RGB}{0,105,252}
\definecolor{myred}{RGB}{247,96,102}
\definecolor{mygray}{gray}{0.4}
\definecolor{red}{RGB}{255, 0, 0}
\definecolor{green}{RGB}{0, 100, 0}
\definecolor{gold}{RGB}{255, 125, 0}

\def\etal{\emph{et al}}

\newcommand{\VarSty}[1]{\textnormal{\ttfamily\color{blue!90!black}#1}\unskip}
\newcommand{\var}[1]{\textcolor{black}{\texttt{#1}}}
\SetKwProg{For}{for}{:}{}

\usepackage{newfloat}
\usepackage{listings}
\DeclareCaptionStyle{ruled}{labelfont=normalfont,labelsep=colon,strut=off} 
\floatstyle{ruled}
\newfloat{listing}{tb}{lst}{}
\floatname{listing}{Listing}
\title{\model: Self-Adaptive Touch-Language Framework for Tactile Commonsense Reasoning in Open-Ended Scenarios}

\author {
    Ning Cheng\textsuperscript{\rm 1,2},
    Jinan Xu\textsuperscript{\rm 1,2},
    Jialing Chen\textsuperscript{\rm 1,2},
    Bin Fang\textsuperscript{\rm 3}, 
    Wenjuan Han\textsuperscript{\rm 1,2}\thanks{Corresponding author. Email address: wjhan@bjtu.edu.cn.}
}
\affiliations {
    \textsuperscript{\rm 1} Key Laboratory of Big Data \& Artificial Intelligence in Transportation (Beijing Jiaotong University), Ministry of Education\\
    \textsuperscript{\rm 2}School of Computer Science and Technology, Beijing Jiaotong University, Beijing 100044, China\\
    \textsuperscript{\rm 3}Beijing University of Posts and Telecommunications, 100876, Beijing, China
}

\usepackage{bibentry}

\begin{document}

\maketitle

\begin{abstract}
This paper explores the challenges of integrating tactile sensing into intelligent systems for multimodal reasoning, particularly in enabling commonsense reasoning about the open-ended physical world. We identify two key challenges: \textbf{modality discrepancy}, where existing touch-language models often treat touch as a mere sub-modality of language without further addressing the semantic differences, and \textbf{open-ended tactile data scarcity}, where current datasets lack the diversity, open-endedness, and complexity needed for reasoning. To overcome these challenges, we introduce \model, a \textbf{S}elf-Adaptive \textbf{To}uch-\textbf{La}nguage framework. \model utilizes Mixture of Experts (MoE) to dynamically process, unify, and manage tactile and language modalities, capturing their unique characteristics. Crucially, we also present a comprehensive tactile commonsense reasoning dataset and benchmark featuring free-form questions and responses, 8 physical properties, 4 interactive characteristics, and diverse commonsense knowledge. Experiments show \model exhibits competitive performance compared to existing models on the \textsc{PhysiCLeAR} benchmark and self-constructed datasets, proving the effectiveness of the Mixture of Experts architecture in multimodal management and the performance advantages for open-scenario tactile commonsense reasoning tasks.
\end{abstract}

\begin{links}
    \link{Project Page}{https://cocacola-lab.github.io/SToLa-Page/}
\end{links}

\section{Introduction}
\label{sec:intro}

Human interactions with the physical world are fundamentally grounded in touch, a sense that surpasses the constraints of vision and hearing, offering direct, detailed, and multidimensional perception through physical contact~\cite{fulkerson2013first,paterson2020senses,packheiser2024systematic}. 
In the field of robotics and artificial intelligence, tactile sensing has been widely recognized as a critical modality for robots to interact with their surroundings, especially in scenarios with visual occlusion~\cite{kappassov2022tactile,lenz2024analysing,ueda2024visuo}.

\begin{figure}[t]
\begin{center}
\includegraphics[width=\linewidth]{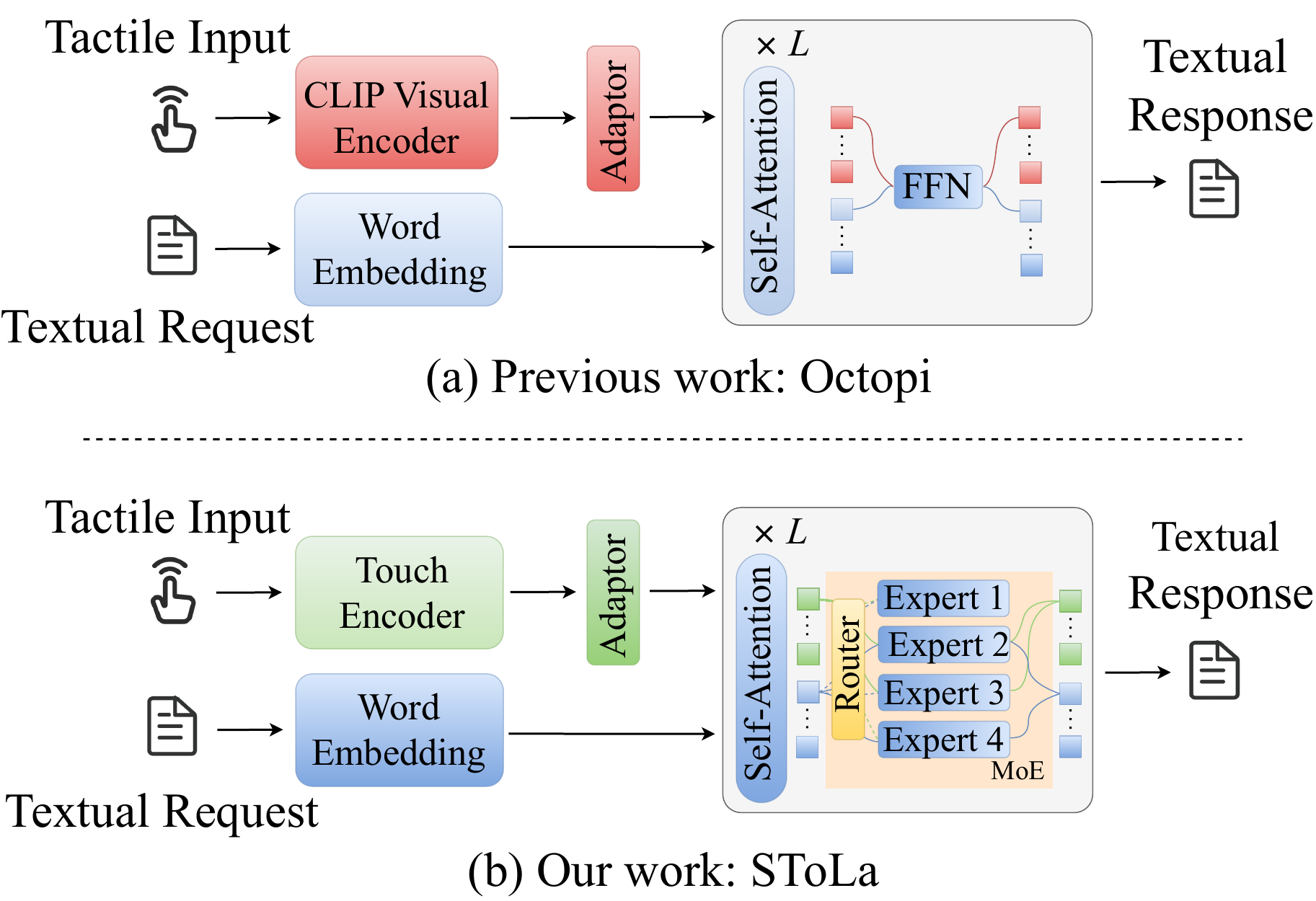}
\caption{Comparison between (a) previous work and (b) our work across model.}
\label{fig:comparison}
\end{center}
\vskip -0.2in
\end{figure}

However, integrating tactile sensing for reasoning presents significant challenges. These challenges can be distilled into two key issues. Firstly, a fundamental \textbf{modality discrepancy} exists: tactile and language modalities possess distinct characteristics, a fact underscored by dedicated neural pathways for touch processing. Current touch-language models often oversimplify this, treating touch as a mere ``sub-modality'' of language without further addressing the semantic differences. That is, these models use a touch encoder to map tactile data into a representation space that's very similar to the representation space used for text, and then force touch and language representation to fit into one transformer architecture~\cite{yu2024octopi,yang2024binding}, overlooking the fact that the two representations remain semantically distinct despite being mapped into a shared space. This lack of further distinction prevents the model from understanding the nuanced differences between the two modalities. Secondly, we face \textbf{open-ended tactile data scarcity}. Natural interaction demands intelligent systems capable of handling free-form queries and responses encompassing a broad spectrum of tactile properties. However, existing datasets, such as the recently recognized \textsc{PhysiCLeAR}~\cite{yu2024octopi}, are limited in scope, focusing on a narrow range of properties 
and employing templated question-answer formats. This constraint severely restricts the generalization capabilities of models in real-world, open-ended scenarios.

To address the challenge of modality discrepancy, we pioneer the exploration of the Mixture of Experts (MoE) to dynamically process diverse token types from both tactile and language modalities within the tactile domain, and propose \textbf{\model} (\textbf{S}elf-Adaptive \textbf{To}uch-\textbf{La}nguage), a framework capable of effectively handling both single tactile images and tactile sequential data, while leveraging MoE to seamlessly modality integration. As illustrated in Figure~\ref{fig:comparison}, \model primarily distinguishes itself from typical touch-language models, such as Octopi~\cite{yu2024octopi}, by incorporating an MoE layer within the internal blocks of the Large Language Model (LLM). Each MoE-enabled block has a shared self-attention layer that is applicable to both tactile and language modalities, alongside the routers and feed-forward network (FFN) based experts to dynamically allocate token-level knowledge for both modalities. To further enhance training stability and expert collaboration, we employ a two-stage progressive training strategy. Through this model architecture and training strategy, we obtained an efficient and stable \model model.

To address the challenge of open-ended tactile data scarcity, we introduce a comprehensive commonsense reasoning dataset. This novel dataset transcends the limitations of existing resources, encompassing over 8 physical properties, 4 interactive characteristics, and diverse commonsense knowledge.
Notably, the dataset features free-form queries and responses, designed to reflect the complexities of general tactile open-ended scenarios. In contrast, the widely recognized \textsc{PhysiCLeAR}~\cite{yu2024octopi} dataset, while valuable, focuses on a limited scope, exploring commonsense reasoning across only three physical properties: hardness, roughness, and bumpiness. Furthermore, despite its five reasoning tasks, \textsc{PhysiCLeAR} relies heavily on templated question-answer formats. For instance, the object property description subtask has only a few question formats, mainly variations of word substitutions, with responses following a fixed structure. The other four tasks are similar. However, in real-world scenarios, the form of both questions and responses is unpredictable, highlighting a significant gap between the current dataset and the demands of genuine open-ended tactile reasoning.

To validate the effectiveness of the \model~framework including the model architecture, training strategy, and data set, we compare it with state-of-the-art touch-language models. Our experimental results demonstrate that dynamically managing different modalities through MoE significantly outperforms the traditional approach. Our contributions are summarized as follows:	
\begin{itemize}
\setlength{\itemsep}{0pt}
\setlength{\parsep}{0pt}
\setlength{\parskip}{0pt}
  \item \textit{Framework.} We propose \model, a pioneering MoE-based touch-language framework capable of processing diverse forms, including individual tactile images and tactile time-series data, as well as accommodating different sensor configurations (GelSight and GelSight Mini). We also present a progressive training paradigm to enhance the LLM's comprehension of the tactile and language modality.
  \item \textit{Dataset.} We introduce a comprehensive tactile commonsense reasoning dataset for open-ended scenarios, featuring free-form questions and answers. The dataset covers over 8 physical properties, 4 interactive characteristics, and various commonsense knowledge from daily life.
  \item \textit{Practice.} \model surpasses existing touch-language models in overall performance on \textsc{PhysiCLeAR} and TactileBench. \model also delivers competitive results across various subtasks. 
\end{itemize}

\section{Related Work}
\label{sec:work}
\subsection{Tactile Commonsense Reasoning}
Existing models process tactile signals by leveraging the powerful reasoning capabilities of LLMs. Yang \etal. \cite{yang2024binding} aligns touch embeddings with image embeddings from the existing vision-language model \cite{zhang2023llama, gao2023llama} through contrastive learning, resulting in the creation of the Touch-LLM, a touch-language model capable of performing tactile question-answering tasks, including tactile commonsense reasoning tasks. Yu \etal.  \cite{yu2024octopi} have significantly advanced tactile commonsense reasoning by formally introducing the \textsc{PhysiCLeAR} dataset, a training and evaluation suite based on three physical properties: hardness, roughness, and bumpiness. This suite includes five tactile commonsense reasoning tasks based on tactile temporal signals, leading to the development of Octopi, a touch-language model based on Vicuna v1.5 \cite{chiang2023vicuna}. Unlike previous work, we focus more on free-form tactile commonsense reasoning in open scenarios that align with real-world distributions.

\subsection{Mixture of Experts}
Mixture of Experts (MoE) aims to boost performance by selectively activating a subset of experts via a routing mechanism, enabling efficient handling of diverse data. Fedus \etal. \cite{fedus2022switch} propose Switch Transformer, Du \etal.  \cite{du2022glam} introduce GLaM, and Komatsuzaki \etal. \cite{komatsuzaki2023sparse} present sparse upcycling method, all of which demonstrate MoE's performance advantages and exceptional efficiency in language models. Additionally, MoE has achieved groundbreaking progress in vision models \cite{riquelme2021scaling,chen2023adamv,chen2024eve,zhu2024moe}. Recently, MoE has been widely applied in multimodal models \cite{lin2024moe,shu2024llava,li2025uni}. In this work, we pioneer the integration of MoE into the large touch-language model, enabling finer differentiation, management, and interpretation across tactile and language modalities. 

\begin{figure*}[htp]
\begin{center}
\centerline{\includegraphics[width=\linewidth]{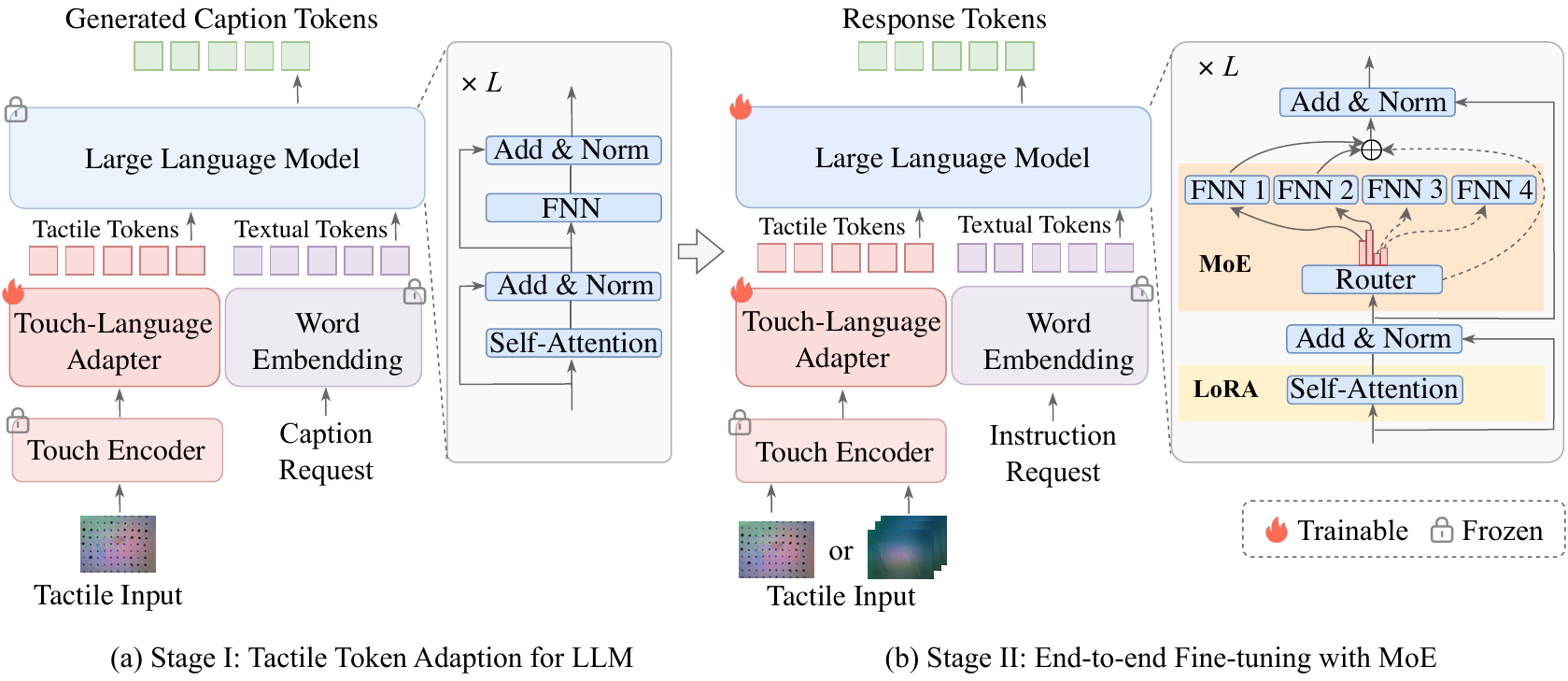}}
\caption{\textbf{\model~framework.} Our framework consists of a touch encoder, a touch-language adapter, and a LLM. The training process follows a two-stage strategy. Stage I: We train only the touch-language adapter, allowing the LLM to adapt to tactile inputs—static tactile images with spatial details. Stage II: The weights from Stage I are copied, keeping the touch encoder unchanged. The self-attention module of the LLM is fine-tuned using LoRA, while the FFN is upcycled from dense to sparse. Notably, we do not adjust the word embedding layer throughout the process.}
\label{fig:model}
\end{center}
\vskip -0.2in
\end{figure*}

\section{Method}
We introduce a \textbf{S}elf-Adaptive \textbf{To}uch-\textbf{La}nguage Framwork (\model) for tactile commonsense reasoning, capable of handling different forms of tactile input, but also manages both tactile and language modalities effectively, as illustrated in Figure~\ref{fig:model}. In this section, we detail the \model's model architecture and the two-stage training strategy.

\subsection{Model Architecture}
\textbf{Overview.}
As shown in Figure~\ref{fig:model}, \model~comprises three key components: a touch encoder, a touch-language adapter, and a LLM with MoE blocks. First, the touch encoder processes raw tactile data, transforming it into corresponding embeddings. Subsequently, the touch-language adapter bridges the modality gap, performing a coarse alignment of these embeddings with textual representations. Finally, and crucially, the LLM itself is augmented with Mixture of Experts (MoE) blocks. 
Next, we elaborate on the model architecture in the following paragraphs.

The \textit{Input Unification} paragraph introduces how the touch encoder unifies diverse tactile signals and how the adapter and LLM integrate embeddings from the tactile and language modalities; \textit{System Design} paragraph introduces the workflow of each component; \textit{MoE Module} paragraph introduces how to apply the MoE module in the touch-language transformer.

\noindent\textbf{Input Unification.}
To enable the self-adaptive touch-language model, it is essential to unify the model's inputs by seamlessly integrating tactile signals with text, harmonizing individual tactile images alongside time-series data, and accommodating diverse sensor configurations, such as Gelsight and Gelsight Mini. Following previous work~\cite{gao2023objectfolder,yang2024binding,higuerasparsh,feng2025anytouch,dave2024multimodal}, tactile signals are divided into static images (individual tactile images) and dynamic videos (time-series data). To harmonize the two forms of tactile inputs, we process images as single-frame videos. Given a tactile video input $X_{touch} \in \mathbb{R}^{N \times H \times W \times C}$ with
$N$ frames, where $H$ and $W$ are the initial resolution of a frame, and $C$ is the number of channels, with $C=3$ in this case. The touch encoder encoders the $N$ frames independently as a batch of tactile images and produces frame-level tactile token sequences $\mathcal{Z}=[[z_{11},z_{12},\cdots,z_{1P}],\cdots,[z_{i1},z_{i2},\cdots,z_{ij},\cdots,z_{1P}], \cdots, \newline [z_{N1},z_{N2},\cdots,z_{NP}]]\in\mathbb{R}^{N\times P\times C}$, where $ij$ represents the $j$-th tactile token in the $i$-th frame, and $P=\frac{H \times W}{14^2}$ denotes the sequence length of tactile tokens. Each tactile token is a $14\times14$ patch. Inspired by ViFi-CLIP \cite{rasheed2023fine}, these frame-level tactile token sequences are average-pooled to obtain a video-level tactile token sequence $\mathcal{Z}^{'}\in\mathbb{R}^{P\times C}$. This operation aggregates multiple frames, implicitly incorporating temporal patterns. It is noteworthy that all sensor configurations, including both Gelsight and Gelsight Mini images, use the same processing method. Subsequently, the touch-language adapter $f_{touch}$ is applied to transform $\mathcal{Z}^{'}\in\mathbb{R}^{P\times C}$ to $\mathcal{V}\in\mathbb{R}^{P\times D}$, with $D$ denoting the hidden size of the LLM. In addition, the text input is processed through a word embedding layer $f_{text}$, which maps the text input to the sequence tokens $\mathcal{T}=[t_{1},t_{2},\cdots,t_{M}]\in\mathbb{R}^{M\times D}$, where $M$ refers to the length of text token sequence.  Finally, we concatenate the tactile tokens and text tokens together and input the resulting sequence into the LLM.

\noindent\textbf{System Design.}
We leverage the existing tactile representation model from TLV-Link~\cite{cheng2024touch100k}, which is well-aligned with the language modality, as our touch encoder. Meanwhile, we use a linear projection layer and Vicuna-7b v1.5~\cite{chiang2023vicuna} to function as our touch-language adapter and LLM, respectively. Given a tactile-textual instruction conversation $(X,Y)$, \model~produces response $Y$ as follows:
\begin{equation}  Y=LLM_\phi(Proj_\lambda(Enc_{\omega}(X_{touch})),X_{text}),
\end{equation}
where $X_{touch}$ is the tactile input, and $X_{text}$ is the text request. $Enc$, $Proj$, and $LLM$ refer to the touch encoder, touch-language adapter, and LLM, respectively, with $\omega$, $\lambda$, and $\phi$ denoting their corresponding parameters.

\noindent\textbf{MoE Module.} 
Considering that the commonsense reasoning task involves diverse tokens from multiple modalities, we introduce MoE layers into the model to dynamically select and activate experts, allowing it to adapt to varying input patterns. As shown in Figure~\ref{fig:model}, our MoE layer consists of a router and multiple experts. Specifically, we use a linear layer as the router and replicate the FFNs from Stage I to form $K$ experts $\{E_{i}\}^{K}_{i=1}$. The router is responsible for predicting the activation probability of each expert for each token $x$, and this process can be formalized as:
\begin{equation}
    \mathcal{P}(\mathbf{x})=Softmax(\text{Top-}{k}(x \cdot W_{r}, k)),
\end{equation}
where $W_r \in \mathbb{R}^{D \times K}$ represents the router’s weight matrix, and $\text{Top-}{k}(x \cdot W_{r}, k)$ selects the top $k$ experts based on the router’s weight logits $x \cdot W_{r} \in \mathbb{R}^{1 \times K}$. The $\text{Top-}{k} $ strategy for expert selection can be defined as:
\begin{equation}
\text{Top-}k_{i}({z,k}) = 
\begin{cases} 
z_i, & \text{if } z_{i} 
\text{ is in the top k values of } z. \\
\infty, & \text{otherwise}.\\
\end{cases}
\end{equation}
Thus, the MoE layer output is calculated as the weighted sum of the experts’ contributions, with the activation probabilities functioning as the weights: 
\begin{equation}
    \mathrm{MoE}(x)=\sum_{i=1}^K\mathcal{P}_i(x)\cdot E_i(x).
\end{equation}

\subsection{Two-Stage Training Strategy}
\label{sec:training_strategy}
\textbf{Stage I: Tactile Token Adaption for LLM.}
The goal at this stage is to adapt the tactile tokens converted by the touch encoder into LLM, enabling the LLM to interpret the content within the tactile inputs. The strategy involves using a touch-language adapter to map the tactile tokens into the LLM’s text representation space, treating tactile patches as \textbf{pseudo}-text tokens. Specifically, the touch-language adapter is trained on tactile images and parallel language descriptions, while the touch encoder and LLM remain frozen. During this stage, the MoE layers are NOT employed to the LLM. We minimize the cross-entropy loss of the generated tokens to optimize the output $\mathcal{Y}=[y_{1},y_{2},\cdots,y_{M}]\in\mathbb{R}^{M\times D}$ of the LLM. The objective function is:
\begin{equation}
\mathcal{L}_{\text{ce}} = -\mathbb{E}_{(\mathcal{Y}_i \mid \mathcal{V}, \mathcal{T}_{<i}) \sim \pi_{\theta}} \left[ \log \pi_{\theta}(\mathcal{Y}_i \mid \mathcal{V}, \mathcal{T}_{<i}) \right],
\end{equation}
where the training process employs teacher forcing \cite{williams1989learning}, and $\pi_{\theta}(\mathcal{Y}_i\mid\mathcal{V}, \mathcal{T}_{<i})$ represents the likelihood of the predicted token $\mathcal{Y}_{i}$ conditioned on $\mathcal{V}$ and the first $i-1$ target tokens $\mathcal{T}_{<i}$.

\noindent\textbf{Stage II: End-to-end Fine-tuning with MoE.} 
In this stage, we aim to dynamically assign specialized experts to process diverse token types from both touch and language modalities, thereby enhancing the model's multimodal comprehension and generative abilities. To achieve this, we freeze only the touch encoder and word embedding layer, while fine-tuning the touch-language adapter and the LLM using instruction data. Within the LLM, we employ LoRA~\cite{hu2022lora} for parameter-efficient fine-tuning of the self-attention layers, and crucially, replace the traditional feed-forward network (FFN) layers with Mixture of Experts (MoE) layers. This combination enables dynamic, token-level expertise allocation, significantly improving the model's ability to handle complex touch-language inputs. 
Specifically, for the MoE layers, a linear layer is employed as the router, with the FFN from stage I being replicated multiple times to facilitate the implementation of the experts. When tactile and textual tokens are fed into the MoE layers, the router calculates the weights of each expert for each token. Using the Top-k strategy, only the top-k experts are activated, and each token is processed by these activated experts. The resulting output is the weighted sum of the router's weights and the outputs from the activated experts. 
In addition to the cross-entropy loss from stage I, we introduce a differentiable load balancing loss~\cite{fedus2022switch} to the MoE layer as an auxiliary loss. The formula is as follows.
\begin{equation}
    \mathcal{L}_{\text{aux}}=\alpha \cdot K\cdot\sum_{i=1}^{K}\mathcal{F}_i\cdot \mathcal{G}_i,
\end{equation} 
where $\alpha$ is the scaling factor, $\mathcal{F}_i$ represent the probability of tokens assigned to expert $E_i$ and $\mathcal{G}_i$ represents the router probability allocated to expert $E_i$, respectively. $\mathcal{F}_i$ and $\mathcal{G}_i$ are calculated using the following formulas:
\begin{equation}
    \mathcal{F}_i=\frac{1}{P+M}\sum_{i=1}^K\mathrm{1}\{\operatorname{argmax}\mathcal{P}(\mathbf{x})=i\},
\end{equation} 
\begin{equation}
    \mathcal{G}_i=\frac{1}{P+M}\sum_{i=1}^{P+M}\mathcal{P}_i(\mathbf{x}).
\end{equation} 

Thus, the objective function for this stage is:

\begin{equation}\mathcal{L}_{\text{total}}=\mathcal{L}_{\text{ce}} + \mathcal{L}_{\text{aux}}.
  \label{eq:gen}
\end{equation}

\begin{table*}
\centering
\begin{small}
\begin{tabular}{lp{1.8cm}p{1.8cm}p{1.8cm}p{1.8cm}p{1.8cm}p{1.8cm}}
\toprule
\multirow{2}{*}{\textbf{Model}} & \multicolumn{3}{c}{\textbf{\textsc{PhysiCLeAR}}}  & \multicolumn{3}{c}{\textbf{TactileBench}} \\ 
\cmidrule(r){2-4} \cmidrule(r){5-7} 
 & \makecell[c]{CIDEr} & \makecell[c]{B@4}  & \makecell[c]{METEOR} & \makecell[c]{METEOR} & \makecell[c]{GPT-4} & \makecell[c]{DeepSeek-R1} \\
\midrule
Touch-LLM \cite{yang2024binding}    & \makecell[c]{-} & \makecell[c]{-} & \makecell[c]{-} & \makecell[c]{17.92} & \makecell[c]{6.88} & \makecell[c]{7.06} \\
Octopi-7B \cite{yu2024octopi}   & \makecell[c]{138.60} & \makecell[c]{64.16} & \makecell[c]{77.63} & \makecell[c]{21.47} & \makecell[c]{6.91} & \makecell[c]{7.17} \\
Octopi-13B \cite{yu2024octopi}   & \makecell[c]{\underline{141.20}} & \makecell[c]{\underline{64.33}} & \makecell[c]{\underline{77.79}} & \makecell[c]{\underline{28.83}} & \makecell[c]{\underline{7.85}} & \makecell[c]{\underline{7.97}} \\
\model (Ours)& \makecell[c]{\textbf{195.03}} & \makecell[c]{\textbf{68.03}} & \makecell[c]{\textbf{82.58}} & \makecell[c]{\textbf{30.27}} & \makecell[c]{\textbf{8.02}} & \makecell[c]{\textbf{8.12}} \\
\bottomrule
\end{tabular}
\end{small}
\caption{\textbf{Overall performance comparison on the \textsc{PhysiCLeAR} and TactileBench benchmark.} The best results
are shown in \textbf{bold}, and the suboptimal ones are highlighted with \underline{underline}. For Touch-LLM, which does not support the \textsc{PhysiCLeAR} dataset with interleaved tactile temporal signals and text, the corresponding results are represented with “–”.}
\label{tab:overall_res}
\end{table*}

\begin{table*}[!ht]
\centering
\begin{small}
\begin{tabular}
{lp{1.2cm}p{1.2cm}p{1.2cm}p{1.2cm}cccc}
\toprule
\multirow{2}{*}{\textbf{ Model}} & \multirow{2}{*}{\textbf{PC}}  & \multirow{2}{*}{\textbf{PSS}} & \multirow{2}{*}{\textbf{POM}} & \multirow{2}{*}{\textbf{PSR}} & \multicolumn{4}{c}{\textbf{OPD}} \\ 
\cmidrule{6-9}
& & & & &Combined &  Hardness &  Roughness &  Bumpiness \\
\midrule

Random &   33.33 &  33.33 &  16.67 &  50.00 & 3.70 & 33.33 & 33.33 & 33.33\\
Octopi-7B \cite{yu2024octopi} &  48.10 &  74.67 &  44.39 &  \underline{69.57} & 47.37 & \underline{71.05} & 73.68 & 81.58 \\
Octopi-7B* \cite{yu2024octopi}&  43.71  & 63.43  & 39.49  & 69.39 & 20.51 & 28.21 & 71.79 & \textbf{92.31} \\
Octopi-13B \cite{yu2024octopi} &  \underline{55.06} &  \textbf{84.00} &  \textbf{60.43} &  67.39 & \textbf{55.26} & \textbf{73.68} & \underline{78.95} & 78.95 \\
Octopi-13B* \cite{yu2024octopi}&  41.92  & 66.29  & 50.32 & 67.35 & 20.53 & 30.77 & 76.92 & \underline{82.05}\\
\model (Ours) &   \textbf{62.28} & \underline{74.86} & \underline{57.32} & \textbf{69.80} & \underline{48.72} & 61.54 & \textbf{82.05} & \underline{82.05} \\
\bottomrule 
\end{tabular}
\end{small}

\captionsetup{justification=raggedright,singlelinecheck=false} 
\caption{\textbf{Subtasks accuracy comparison with accuracy score on the \textsc{PhysiCLeAR} benchmark. }The results marked with * are the ones reproduced by us using the open-source models and scripts provided in the original paper. The best results are in \textbf{bold}, and the second-best ones are \underline{underlined}. \textbf{PC}: Property Comparison. \textbf{PSS}: Property Superlative Selection. \textbf{POM}: Property-object Matching. \textbf{PSR}: Property Scenario Reasoning. \textbf{OPD}: Object Property Description. Combined: The generated results for hardness, roughness, and bumpiness are all correct.}
\label{tab:subtask_res_phys}
\end{table*}

\begin{table*}[!ht]
\centering
\resizebox{\textwidth}{!}{
\begin{small}
\begin{tabular}
{lccccccccc}
\toprule
\multirow{2}{*}{\textbf{ Model}} & \multicolumn{3}{c}{\textbf{FPU}}  & \multicolumn{3}{c}{\textbf{TIP}} & \multicolumn{3}{c}{\textbf{CDR}} \\ 
\cmidrule(r){2-4} \cmidrule(r){5-7} \cmidrule{8-10}
& \footnotesize{METEOR} & \footnotesize{GPT-4} & \footnotesize
{DeepSeek-R1} &  \footnotesize{METEOR} & \footnotesize{GPT-4} &  \footnotesize{DeepSeek-R1} &  \footnotesize{METEOR} & \footnotesize{GPT-4} & \footnotesize{DeepSeek-R1} \\
\midrule
Touch-LLM \cite{yang2024binding} &  15.49 &  7.01 & 7.16  &  17.27 & 6.42 & 6.51 & 24.98 & 7.24 & 7.67 \\
Octopi-7B \cite{yu2024octopi} &  21.87 &  6.65 &  7.04 &  22.55 & 7.13 & 7.15 & 18.82 & 7.26 & 7.52 \\
Octopi-13B \cite{yu2024octopi} & \underline{29.70}  &  \underline{7.81} &  \underline{7.96} &  29.89 &  7.81 & \underline{7.87} & \underline{25.04} & \textbf{8.00} & \textbf{8.15}\\
\model (Ours) &   \textbf{31.34} & \textbf{8.19} & \textbf{8.28} & \textbf{31.24} & \textbf{8.03} & \textbf{7.97} & \textbf{26.15} & \underline{7.61} & \underline{7.96} \\
\bottomrule 
\end{tabular}
\end{small}
}
\caption{\textbf{Subtasks performance comparison on the TactileBench.} The best results are in \textbf{bold}, and the second-best ones are \underline{underlined}. \textbf{FPU}: Fundamental property Understanding. \textbf{TIP}: Tactile Interaction Perception. \textbf{CDR}: Commonsense-Driven Reasoning.} 
\label{tab:subtask_res_tacti}

\end{table*}

\section{TactileBench}

To evaluate models' reasoning capabilities in open-ended unstructured environments, we designed a comprehensive benchmark encompassing tasks ranging from fundamental understanding to complex reasoning. Unlike \textsc{PhysiCLeAR}, which was designed around specific task objectives with templated questions and answers, our benchmark dataset focuses on the hierarchical cognitive process from perception to reasoning. Rather than following a fixed format, we adopt an open-ended approach that better evaluates the model’s depth of understanding and complexity.

\subsection{Task Category}
The underlying tasks involve open-ended responses in a free-form manner. The benchmark consists of three progressively challenging subtasks: 

\noindent\textbf{Fundamental Property Understanding.}
The subtask involves recognizing and describing an object’s basic physical properties, including but not limited to hardness, roughness, weight, and texture. The model needs to perceive these properties through tactile signals and convert them into human-understandable textual outputs.

\noindent\textbf{Tactile Interaction Perception.} The subtask involves sensing an object’s dynamic characteristics during real-world interactions. These characteristics include graspability, prickliness, bendability, malleability, and more. The model needs to perceive and respond to these complex tactile signals in real time during dynamic interactions with objects.

\noindent\textbf{Commonsense-Driven Reasoning.} The subtask requires the model not only to perceive object’s tactile properties but also to integrate external commonsense knowledge for reasoning. These tasks often involve understanding an object’s behavior, function, or usage in specific scenarios, as well as making high-level decisions based on tactile information.

\subsection{Data Construction}
We use the material classification test set from \textit{Touch and Go} \cite{yang2022touch} as our baseline data. This test set consists of visual images, tactile images, and classification labels. To construct tactile question-answering data for three types of tasks, we designed a unified prompt template using visual images and classification labels, queried GPT-4, and conducted manual verification.

\subsection{Data Statistics}

Given the significance of fundamental property understanding, the auxiliary role in tactile interaction perception, and the challenges of commonsense-driven reasoning, the data are stratified into a 50\%-30\%-20\% proportion to ensure a balanced difficulty distribution. Ultimately, we compile a total of 600 questions, each with 3-5 ground-truth answers, covering 14 objects.

\subsection{Evaluation Metrics}

Since TactileBench is an open-ended, free-form tactile commonsense reasoning dataset, traditional n-gram-based metrics such as CIDEr \cite{vedantam2015cider} and B@4 \cite{papineni2002bleu} are not suitable. Instead, we evaluate the dataset using METEOR \cite{banerjee2005meteor}, which is based on semantic similarity, along with multi-dimensional scoring from GPT-4 \cite{achiam2023gpt} and DeepSeek–R1 \cite{guo2025deepseek}. 

\begin{figure}[t]
\begin{center}
\centerline{\includegraphics[width=\linewidth]{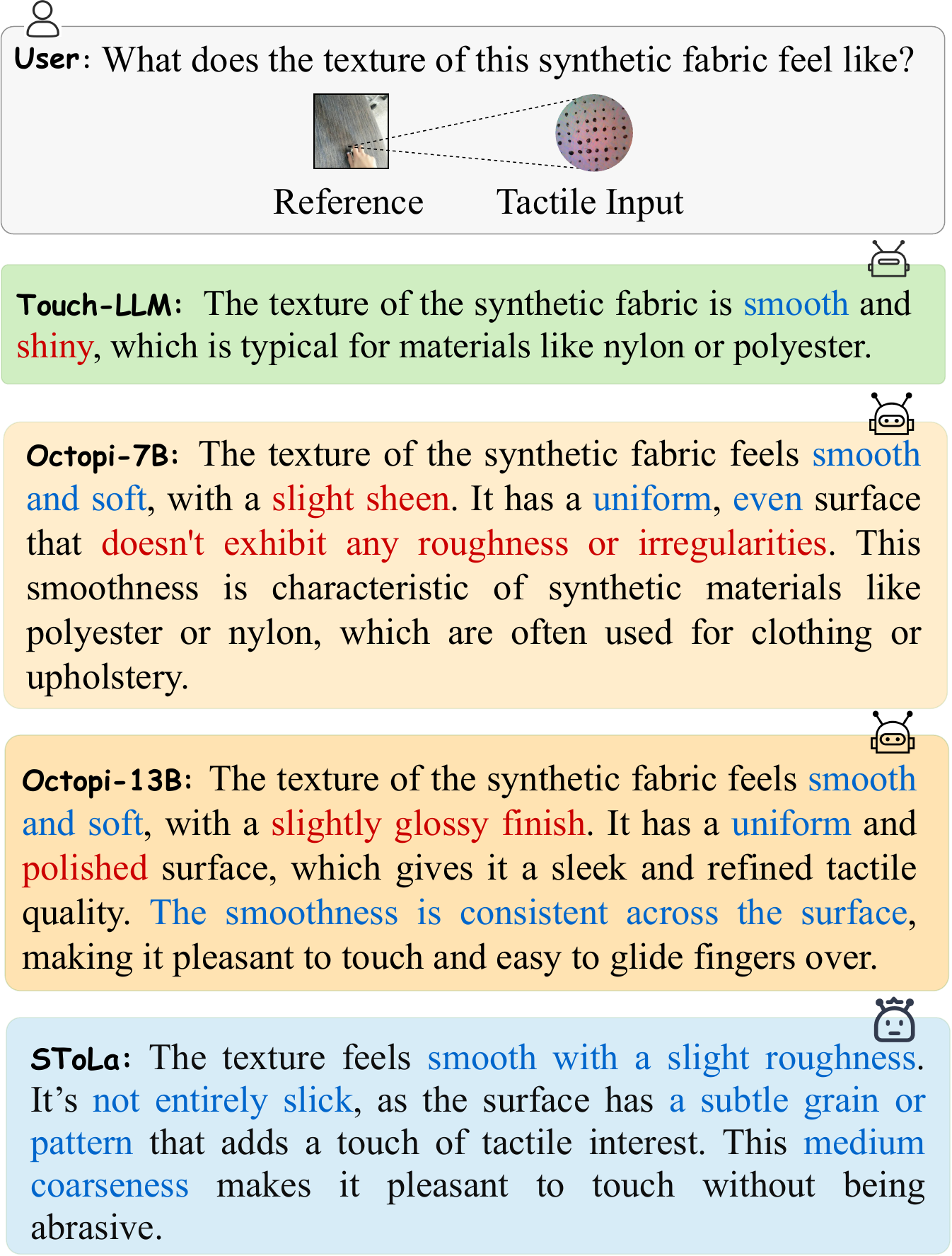}}
\caption{\textbf{Qualitative responses of \model and existing touch-language models.} Errors are highlighted in \textcolor{red}{red}, while detailed and accurate parts are emphasized in \textcolor{myblue}{blue}. For visualization, we include the visual signals of the tactile input (labeled as “reference”), even though it is not used for the model.}
\label{fig:more_case}
\end{center}
\vskip -0.2in
\end{figure}

\begin{figure}[!t]
\begin{center}
\includegraphics[width=\linewidth]{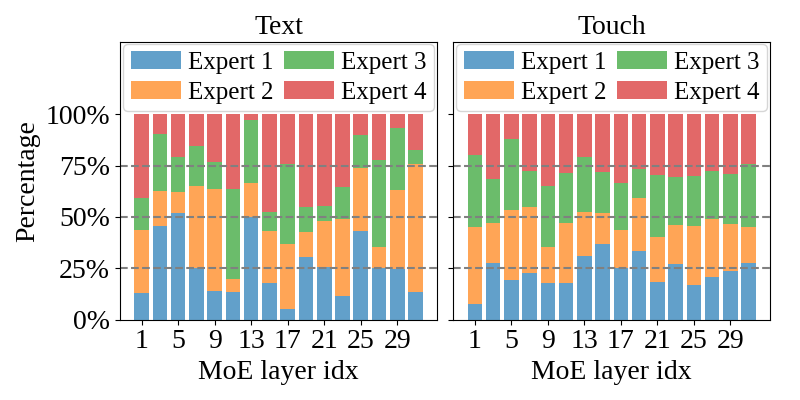}
\caption{Distribution of modalities across different experts.}
\label{fig:vis2}
\end{center}
\end{figure}

\section{Experiments}

\subsection{Implementation Details}
Models are trained with batch size 16 on an Nvidia A100-80G GPU. During the implementation of our two-stage training strategy, we treat Stage I as a ``Touch-to-Text'' generation task, trained on touch-language pairs, with the objective being to prompt the LLM to generate a corresponding text description given a tactile input. Specifically, we use touch-language pairs from Touch100k. Stage II is regarded as a process of instruction tuning, aimed at enhancing the model's capabilities and controllability. In this stage, we use the video-based \textsc{PhysiCLeAR} dataset and our self-constructed image-based tactile instruction dataset. Both datasets are processed into video units by frames. 
In particular, the base model of \model adopts Vicuna-7B v1.5. 

\begin{figure}[]
\begin{center}
\includegraphics[width=\linewidth]{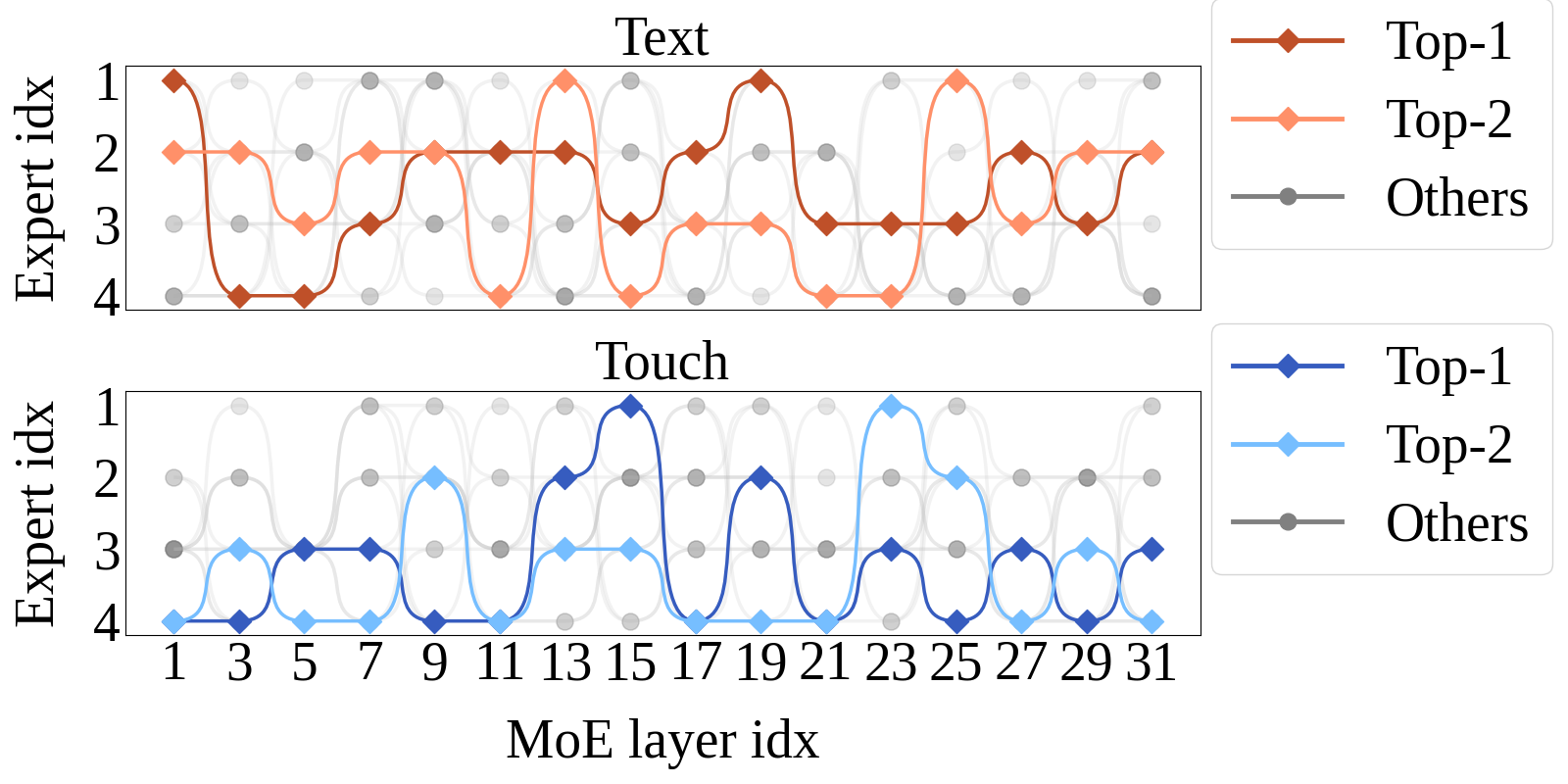}
\caption{Visualization of activated pathways.}
\label{fig:vis3}
\end{center}
\end{figure}

\begin{table*}[!ht]
\centering
\begin{small}
\begin{tabular}
{lp{2.1cm}p{2.1cm}p{2.1cm}p{2.1cm}p{2.1cm}p{2.1cm}}
\toprule
\multirow{2}{*}{\textbf{ Model}} & \multicolumn{3}{c}{\textbf{\textsc{PhysiCLeAR}}}  & \multicolumn{3}{c}{\textbf{TactileBench}} \\ 
\cmidrule{2-7}
& \makecell[c]{CIDEr} & \makecell[c]{B@4}  & \makecell[c]{METEOR} &  \makecell[c]{METEOR} &  \makecell[c]{GPT-4} & \makecell[c]{DeepSeek-R1}  \\
\midrule
\model &   \makecell[c]{\textbf{195.03}} & \makecell[c]{\textbf{68.03}} & \makecell[c]{\textbf{82.58}} & \makecell[c]{\textbf{30.27}} & \makecell[c]{\textbf{8.02}} & \makecell[c]{\textbf{8.12}} \\
\quad w/o MoE &  \makecell[c]{176.79} &  \makecell[c]{66.46} &  \makecell[c]{81.55} &  \makecell[c]{28.71} & \makecell[c]{7.44} & \makecell[c]{7.57}  \\
\quad w/o LoRA &  \makecell[c]{166.71}  & \makecell[c]{64.46}  & \makecell[c]{80.39}  & \makecell[c]{29.32} & \makecell[c]{7.95} & \makecell[c]{7.97} \\
\quad w/o Stage I &  \makecell[c]{172.52} &  \makecell[c]{64.47} &  \makecell[c]{80.55} &  \makecell[c]{29.27} & \makecell[c]{7.72} & \makecell[c]{7.89} \\
\bottomrule 
\end{tabular}
\end{small}
\caption{Ablation study on \textsc{PhysiCLeAR} and TactileBench.}
\label{tab:ablation}
\end{table*}

\subsection{Results}

We conducted a quantitative
 comparison between \model and current state-of-the-art touch-language models, including Touch-LLM \cite{yang2024binding}, Octopi-7B \cite{yu2024octopi}, and Octopi-13B \cite{yu2024octopi}. Table~\ref{tab:overall_res} presents the overall performance on the \textsc{PhysiCLeAR} and TactileBench datasets, while Table~\ref{tab:subtask_res_phys}\footnote{Since the random seed in open-source scripts used for processing the raw data is unknown, our test data differs from that of Octopi \cite{yu2024octopi}. We speculate that the split test sets are more challenging, which results in lower reproduction results compared to those reported in the paper.} and Table~\ref{tab:subtask_res_tacti} provide results comparisons for the subtasks in each dataset, respectively.

Compared to Touch-LLM and Octopi-7B, which use the equivalent scale of 7B, \model achieves the \textbf{best} overall performance in both \textsc{PhysiCLeAR} and TactileBench datasets, as well as across all eight subtasks within these datasets. Since Touch-LLM does not support data in which tactile temporal signals and text appear interleaved, we did not evaluate Touch-LLM’s performance on the \textsc{PhysiCLeAR} dataset. 

Moreover, \model of 7B outperforms Octopi-13B in overall performance and most subtasks on the \textsc{PhysiCLeAR} dataset. Specifically, \model outperforms the previous best-performing Octopi-13B by 53.83, 3.70, and 4.79 in the overall performance metrics of CIDEr, B@4, and METEOR, respectively. Additionally, \model shows superior performance on the property comparison and property scenario reasoning subtasks while achieving suboptimal results only to Octopi-13B on the property superlative selection, property-object matching, and object property description subtasks, demonstrating the strong competitive capabilities of \model with a 7B LLM. It is important to note that we follow the evaluation of Octopi \cite{yu2024octopi} and adopt accuracy as the metric for subtasks. That is, based on templated responses, only the conclusion part of the response is evaluated using exact matching, without considering the reasoning process or semantics. 

Similarly, on the TactileBench dataset, \model of 7B outperforms Octopi-13B in overall performance, whether measured by the METEOR metric or the  GPT-4 and DeepSeek-R1 evaluations. Although Octopi-13B performed slightly under on the commonsense-driven reasoning subtask, \model demonstrates significant advantages in achieving such remarkable results with a significantly smaller parameter size (7B\textless13B). 

Furthermore, we present a qualitative comparison of responses from different models in Figure~\ref{fig:more_case}. The comparison shows that our model excels in interpreting tactile signals and performing tactile reasoning, further confirming the advantages of our approach. More quantitative comparisons can be found in the extended version.

\subsection{MoE Analysis}

We analyze how different experts in each MoE layer of \model dynamically manage the tactile and text modalities. Specifically, the routing distributions and token pathways of \model on TactileBench are visualized in Figure~\ref{fig:vis2} and \ref{fig:vis3}.

\noindent\textbf{Routing Distributions.}
Figure~\ref{fig:vis2} illustrates the modality distribution handled by different experts, revealing that each expert develops its preferences. Differences in the selection of tactile and text modalities indicate that our model can dynamically adjust expert utilization based on input characteristics, enabling effective processing of multimodal data.

\noindent\textbf{Token Pathways.}
We track the trajectories of all tokens and analyze expert behavior at the token level, as shown in Figure~\ref{fig:vis3}. Following the previous work \cite{li2025uni,lin2024moe}, we employ PCA \cite{pearson1901liii} to extract the top 10 pathways for all activated routes. The token pathways also reflect the preferences of different experts for tactile and text modalities across each MoE layer. For a tactile token, \model tends to assign it to experts 3 and 4. Meanwhile, for a text token, \model usually prefers to assign it to a combination of expert 2 and another expert in the shallow layers, while in the deeper layers, expert 3, along with another expert, tends to be assigned. This indicates that \model has learned a specific pattern that enables it to manage the tactile and text modalities in a certain manner.

\subsection{Ablation Study}
In this section, we perform ablation studies to analyze the impact of key factors in \model model, including the MoE module, LoRA, and training strategy, as shown in Table~\ref{tab:ablation}.

\noindent\textbf{Impact of MoE Module.}
To evaluate the MoE module's contribution, we compare models' performance with and without MoE. Specifically, we replace the MoE layer with a standard feedforward network while keeping all other configurations unchanged. The results show that removing the MoE module significantly drops model performance, indicating that the expert routing mechanism plays a crucial role in enhancing performance. 

\noindent\textbf{Impact of LoRA Fine-tuning.} To investigate the impact of LoRA, we ablate low-rank adaptation based on our training processing. The results in Table~\ref{tab:ablation} indicate that LoRA fine-tuning has a significant positive effect on model performance. Without LoRA fine-tuning, the model’s performance degrades significantly.

\noindent\textbf{Impact of Training Strategy.} We further validate the necessity of Stage I in the training strategy by removing it and directly applying the training of Stage II. The experimental results demonstrate that skipping Stage I leads to a decrease in model performance, which proves that the progressive training strategy is crucial to gradually optimizing model capability.

\section{Conclusion}
This work introduces \model, a self-adaptive touch-language framework that pushes the boundaries of tactile commonsense reasoning. By incorporating MoE and two-stage training strategy, \model effectively bridges the gap between tactile and language modalities, enabling more complex reasoning of the open-ended world. Moreover, we contribute a comprehensive tactile commonsense reasoning dataset that covers a wider range of dimensions, with a distribution that aligns with cognitive levels and features free-form content. Finally, experiments demonstrate that our model achieves outstanding performance in tactile commonsense reasoning in open-ended scenarios.

Although \model performs competitive capabilities, computational resource constraints hinder the extension of our method to a 13B LLM, leading to suboptimal performance in certain tasks compared to Octopi-13B. Additionally, we designed the MoE architecture from the modality level. In future work, we will consider assigning experts from a task perspective or a combination of modality and task perspective.

\section*{Acknowledgements}
The work described in this paper has been supported by Fundamental Research Funds for the Central Universities under Grant No. 2025JBZX058, and by National Natural Science Foundation of China under Grant No. 62376019, 62476023, 62406020, 62573063, 62536001.

\bibliography{main}

@article{zhu2024moe,
  title={Moe jetpack: From dense checkpoints to adaptive mixture of experts for vision tasks},
  author={Zhu, Xingkui and Guan, Yiran and Liang, Dingkang and Chen, Yuchao and Liu, Yuliang and Bai, Xiang},
  journal={Advances in Neural Information Processing Systems},
  volume={37},
  pages={12094--12118},
  year={2024}
}

@inproceedings{chen2024eve,
  title={EVE: efficient vision-language pre-training with masked prediction and modality-aware moe},
  author={Chen, Junyi and Guo, Longteng and Sun, Jia and Shao, Shuai and Yuan, Zehuan and Lin, Liang and Zhang, Dongyu},
  booktitle={Proceedings of the AAAI Conference on Artificial Intelligence},
  volume={38},
  number={2},
  pages={1110--1119},
  year={2024}
}

@inproceedings{dave2024multimodal,
  title={Multimodal visual-tactile representation learning through self-supervised contrastive pre-training},
  author={Dave, Vedant and Lygerakis, Fotios and Rueckert, Elmar},
  booktitle={2024 IEEE International Conference on Robotics and Automation (ICRA)},
  pages={8013--8020},
  year={2024},
  organization={IEEE}
}

@inproceedings{banerjee2005meteor,
  title={METEOR: An automatic metric for MT evaluation with improved correlation with human judgments},
  author={Banerjee, Satanjeev and Lavie, Alon},
  booktitle={Proceedings of the acl workshop on intrinsic and extrinsic evaluation measures for machine translation and/or summarization},
  pages={65--72},
  year={2005}
}

@inproceedings{papineni2002bleu,
  title={Bleu: a method for automatic evaluation of machine translation},
  author={Papineni, Kishore and Roukos, Salim and Ward, Todd and Zhu, Wei-Jing},
  booktitle={Proceedings of the 40th annual meeting of the Association for Computational Linguistics},
  pages={311--318},
  year={2002}
}

@inproceedings{vedantam2015cider,
  title={Cider: Consensus-based image description evaluation},
  author={Vedantam, Ramakrishna and Lawrence Zitnick, C and Parikh, Devi},
  booktitle={Proceedings of the IEEE conference on computer vision and pattern recognition},
  pages={4566--4575},
  year={2015}
}

@inproceedings{yang2022touch,
  title={Touch and go: learning from human-collected vision and touch},
  author={Yang, Fengyu and Ma, Chenyang and Zhang, Jiacheng and Zhu, Jing and Yuan, Wenzhen and Owens, Andrew},
  booktitle={Proceedings of the 36th International Conference on Neural Information Processing Systems},
  pages={8081--8103},
  year={2022}
}

@article{williams1989learning,
  title={A learning algorithm for continually running fully recurrent neural networks},
  author={Williams, Ronald J and Zipser, David},
  journal={Neural computation},
  volume={1},
  number={2},
  pages={270--280},
  year={1989},
  publisher={MIT Press One Rogers Street, Cambridge, MA 02142-1209, USA journals-info~…}
}

@article{shu2024llava,
  title={LLaVA-MoD: Making LLaVA Tiny via MoE Knowledge Distillation},
  author={Shu, Fangxun and Liao, Yue and Zhuo, Le and Xu, Chenning and Zhang, Guanghao and Shi, Haonan and Chen, Long and Zhong, Tao and He, Wanggui and Fu, Siming and others},
  journal={CoRR},
  year={2024}
}

@inproceedings{chen2023adamv,
  title={Adamv-moe: Adaptive multi-task vision mixture-of-experts},
  author={Chen, Tianlong and Chen, Xuxi and Du, Xianzhi and Rashwan, Abdullah and Yang, Fan and Chen, Huizhong and Wang, Zhangyang and Li, Yeqing},
  booktitle={Proceedings of the IEEE/CVF International Conference on Computer Vision},
  pages={17346--17357},
  year={2023}
}

@article{riquelme2021scaling,
  title={Scaling vision with sparse mixture of experts},
  author={Riquelme, Carlos and Puigcerver, Joan and Mustafa, Basil and Neumann, Maxim and Jenatton, Rodolphe and Susano Pinto, Andr{\'e} and Keysers, Daniel and Houlsby, Neil},
  journal={Advances in Neural Information Processing Systems},
  volume={34},
  pages={8583--8595},
  year={2021}
}

@inproceedings{komatsuzaki2023sparse,
  title={Sparse Upcycling: Training Mixture-of-Experts from Dense Checkpoints},
  author={Komatsuzaki, Aran and Puigcerver, Joan and Lee-Thorp, James and Ruiz, Carlos Riquelme and Mustafa, Basil and Ainslie, Joshua and Tay, Yi and Dehghani, Mostafa and Houlsby, Neil},
  booktitle={The Eleventh International Conference on Learning Representations},
year={2023}
}

@inproceedings{higuerasparsh,
  title={Sparsh: Self-supervised touch representations for vision-based tactile sensing},
  author={Higuera, Carolina and Sharma, Akash and Bodduluri, Chaithanya Krishna and Fan, Taosha and Lancaster, Patrick and Kalakrishnan, Mrinal and Kaess, Michael and Boots, Byron and Lambeta, Mike and Wu, Tingfan and others},
  booktitle={8th Annual Conference on Robot Learning},
year={2024}
}

@inproceedings{yang2024binding,
  title={Binding touch to everything: Learning unified multimodal tactile representations},
  author={Yang, Fengyu and Feng, Chao and Chen, Ziyang and Park, Hyoungseob and Wang, Daniel and Dou, Yiming and Zeng, Ziyao and Chen, Xien and Gangopadhyay, Rit and Owens, Andrew and others},
  booktitle={Proceedings of the IEEE/CVF Conference on Computer Vision and Pattern Recognition},
  pages={26340--26353},
  year={2024}
}

@inproceedings{gao2023objectfolder,
  title={The objectfolder benchmark: Multisensory learning with neural and real objects},
  author={Gao, Ruohan and Dou, Yiming and Li, Hao and Agarwal, Tanmay and Bohg, Jeannette and Li, Yunzhu and Fei-Fei, Li and Wu, Jiajun},
  booktitle={Proceedings of the IEEE/CVF Conference on Computer Vision and Pattern Recognition},
  pages={17276--17286},
  year={2023}
}

@article{achiam2023gpt,
  title={Gpt-4 technical report},
  author={Achiam, Josh and Adler, Steven and Agarwal, Sandhini and Ahmad, Lama and Akkaya, Ilge and Aleman, Florencia Leoni and Almeida, Diogo and Altenschmidt, Janko and Altman, Sam and Anadkat, Shyamal and others},
  journal={arXiv preprint arXiv:2303.08774},
  year={2023}
}

@article{packheiser2024systematic,
  title={A systematic review and multivariate meta-analysis of the physical and mental health benefits of touch interventions},
  author={Packheiser, Julian and Hartmann, Helena and Fredriksen, Kelly and Gazzola, Valeria and Keysers, Christian and Michon, Fr{\'e}d{\'e}ric},
  journal={Nature Human Behaviour},
  pages={1--20},
  year={2024},
  publisher={Nature Publishing Group UK London}
}

@book{fulkerson2013first,
  title={The first sense: A philosophical study of human touch},
  author={Fulkerson, Matthew},
  year={2013},
  publisher={MIT press}
}

@book{paterson2020senses,
  title={The senses of touch: Haptics, affects and technologies},
  author={Paterson, Mark},
  year={2020},
  publisher={Routledge}
}

@inproceedings{zhu2024minigpt,
  title={MiniGPT-4: Enhancing Vision-Language Understanding with Advanced Large Language Models},
  author={Zhu, Deyao and Chen, Jun and Shen, Xiaoqian and Li, Xiang and Elhoseiny, Mohamed},
  booktitle={The Twelfth International Conference on Learning Representations},
year={2024}
}

@article{liu2024visual,
  title={Visual instruction tuning},
  author={Liu, Haotian and Li, Chunyuan and Wu, Qingyang and Lee, Yong Jae},
  journal={Advances in neural information processing systems},
  volume={36},
  year={2024}
}

@inproceedings{wu2024next,
  title={NExT-GPT: Any-to-Any Multimodal LLM},
  author={Wu, Shengqiong and Fei, Hao and Qu, Leigang and Ji, Wei and Chua, Tat-Seng},
  booktitle={Forty-first International Conference on Machine Learning},
year={2024}
}

@inproceedings{chen2025sharegpt4v,
  title={Sharegpt4v: Improving large multi-modal models with better captions},
  author={Chen, Lin and Li, Jinsong and Dong, Xiaoyi and Zhang, Pan and He, Conghui and Wang, Jiaqi and Zhao, Feng and Lin, Dahua},
  booktitle={European Conference on Computer Vision},
  pages={370--387},
  year={2025},
  organization={Springer}
}

@article{cheng2024touch100k,
  title={Touch100k: A Large-Scale Touch-Language-Vision Dataset for Touch-Centric Multimodal Representation},
  author={Cheng, Ning and Guan, Changhao and Gao, Jing and Wang, Weihao and Li, You and Meng, Fandong and Zhou, Jie and Fang, Bin and Xu, Jinan and Han, Wenjuan},
  journal={arXiv preprint arXiv:2406.03813},
  year={2024}
}

@inproceedings{yu2024octopi,
  title={Octopi: Object Property Reasoning with Large Tactile-Language Models},
  author={Yu, Samson and Lin, Kelvin and Xiao, Anxing and Duan, Jiafei and Soh, Harold},
  booktitle={Robotics: science and systems},
  year={2024}
}

@article{chiang2023vicuna,
  title={Vicuna: An open-source chatbot impressing gpt-4 with 90\%* chatgpt quality},
  author={Chiang, Wei-Lin and Li, Zhuohan and Lin, Zi and Sheng, Ying and Wu, Zhanghao and Zhang, Hao and Zheng, Lianmin and Zhuang, Siyuan and Zhuang, Yonghao and Gonzalez, Joseph E and others},
  journal={See https://vicuna. lmsys. org (accessed 14 April 2023)},
  volume={2},
  number={3},
  pages={6},
  year={2023}
}

@inproceedings{hu2022lora,
  title={LoRA: Low-Rank Adaptation of Large Language Models},
  author={Hu, Edward J and Wallis, Phillip and Allen-Zhu, Zeyuan and Li, Yuanzhi and Wang, Shean and Wang, Lu and Chen, Weizhu and others},
  booktitle={International Conference on Learning Representations},
year={2022}
}

@article{fedus2022switch,
  title={Switch transformers: Scaling to trillion parameter models with simple and efficient sparsity},
  author={Fedus, William and Zoph, Barret and Shazeer, Noam},
  journal={Journal of Machine Learning Research},
  volume={23},
  number={120},
  pages={1--39},
  year={2022}
}

@article{lin2024moe,
  title={Moe-llava: Mixture of experts for large vision-language models},
  author={Lin, Bin and Tang, Zhenyu and Ye, Yang and Cui, Jiaxi and Zhu, Bin and Jin, Peng and Huang, Jinfa and Zhang, Junwu and Pang, Yatian and Ning, Munan and others},
  journal={arXiv preprint arXiv:2401.15947},
  year={2024}
}

@article{zhang2023llama,
  title={Llama-adapter: Efficient fine-tuning of language models with zero-init attention},
  author={Zhang, Renrui and Han, Jiaming and Liu, Chris and Gao, Peng and Zhou, Aojun and Hu, Xiangfei and Yan, Shilin and Lu, Pan and Li, Hongsheng and Qiao, Yu},
  journal={arXiv preprint arXiv:2303.16199},
  year={2023}
}

@article{gao2023llama,
  title={Llama-adapter v2: Parameter-efficient visual instruction model},
  author={Gao, Peng and Han, Jiaming and Zhang, Renrui and Lin, Ziyi and Geng, Shijie and Zhou, Aojun and Zhang, Wei and Lu, Pan and He, Conghui and Yue, Xiangyu and others},
  journal={arXiv preprint arXiv:2304.15010},
  year={2023}
}

@article{guo2025deepseek,
  title={Deepseek-r1: Incentivizing reasoning capability in llms via reinforcement learning},
  author={Guo, Daya and Yang, Dejian and Zhang, Haowei and Song, Junxiao and Zhang, Ruoyu and Xu, Runxin and Zhu, Qihao and Ma, Shirong and Wang, Peiyi and Bi, Xiao and others},
  journal={arXiv preprint arXiv:2501.12948},
  year={2025}
}

@inproceedings{du2022glam,
  title={Glam: Efficient scaling of language models with mixture-of-experts},
  author={Du, Nan and Huang, Yanping and Dai, Andrew M and Tong, Simon and Lepikhin, Dmitry and Xu, Yuanzhong and Krikun, Maxim and Zhou, Yanqi and Yu, Adams Wei and Firat, Orhan and others},
  booktitle={International conference on machine learning},
  pages={5547--5569},
  year={2022},
  organization={PMLR}
}

@article{li2025uni,
  title={Uni-moe: Scaling unified multimodal llms with mixture of experts},
  author={Li, Yunxin and Jiang, Shenyuan and Hu, Baotian and Wang, Longyue and Zhong, Wanqi and Luo, Wenhan and Ma, Lin and Zhang, Min},
  journal={IEEE Transactions on Pattern Analysis and Machine Intelligence},
  year={2025},
  publisher={IEEE}
}

@article{pearson1901liii,
  title={LIII. On lines and planes of closest fit to systems of points in space},
  author={Pearson, Karl},
  journal={The London, Edinburgh, and Dublin philosophical magazine and journal of science},
  volume={2},
  number={11},
  pages={559--572},
  year={1901},
  publisher={Taylor \& Francis}
}

@inproceedings{feng2025anytouch,
  title={AnyTouch: Learning Unified Static-Dynamic Representation across Multiple Visuo-tactile Sensors},
  author={Feng, Ruoxuan and Hu, Jiangyu and Xia, Wenke and Shen, Ao and Sun, Yuhao and Fang, Bin and Hu, Di and others},
  booktitle={The Thirteenth International Conference on Learning Representations},
year={2025}
}

@inproceedings{lenz2024analysing,
  title={Analysing the Interplay of Vision and Touch for Dexterous Insertion Tasks},
  author={Lenz, Janis and Gruner, Theo and Palenicek, Daniel and Schneider, Tim and Pfenning, Inga and Peters, Jan},
  booktitle={CoRL Workshop on Learning Robot Fine and Dexterous Manipulation: Perception and Control},
year={2024}
}

@inproceedings{ueda2024visuo,
  title={Visuo-Tactile Zero-Shot Object Recognition with Vision-Language Model},
  author={Ueda, Shiori and Hashimoto, Atsushi and Hamaya, Masashi and Tanaka, Kazutoshi and Saito, Hideo},
  booktitle={2024 IEEE/RSJ International Conference on Intelligent Robots and Systems (IROS)},
  pages={7243--7250},
  year={2024},
  organization={IEEE}
}

@article{kappassov2022tactile,
  title={Tactile-based task definition through edge contact formation setpoints for object exploration and manipulation},
  author={Kappassov, Zhanat and Ramon, Juan Antonio Corrales and Perdereau, V{\'e}ronique},
  journal={IEEE Robotics and Automation Letters},
  volume={7},
  number={2},
  pages={5007--5014},
  year={2022},
  publisher={IEEE}
}

@inproceedings{rasheed2023fine,
  title={Fine-tuned clip models are efficient video learners},
  author={Rasheed, Hanoona and Khattak, Muhammad Uzair and Maaz, Muhammad and Khan, Salman and Khan, Fahad Shahbaz},
  booktitle={Proceedings of the IEEE/CVF conference on computer vision and pattern recognition},
  pages={6545--6554},
  year={2023}
}

\clearpage
\appendix
\section{Appendix}
\subsection{Tactile Instruction Data}
To address the challenge of generating open-ended tactile data, we adopt a diverse and cost-effective data generation approach inspired by advancements in visual instruction data creation~\cite{zhu2024minigpt,liu2024visual,wu2024next,chen2025sharegpt4v}. 

To ensure cost-effectiveness, we adopt an automated approach. We utilize GPT-4~\cite{achiam2023gpt} to generate instruction-following data from existing touch-language-vision data pairs. 

To maintain diversity, we generate instruction-following data by extending the Touch100k dataset~\cite{cheng2024touch100k} from two complementary perspectives: the object's intrinsic physical properties, as well as its interactive and perceptual characteristics. Subsequently, we select 8 common physical properties—material, elasticity, roughness, mass, hardness, sharpness, texture, and bumpiness—and 4 interactive characteristics related to interaction and perception: graspability, prickliness, bendability, and malleability.

\begin{table*}[!htbp]\centering
\begin{minipage}{1.0\linewidth}\vspace{0mm}    \centering
\begin{tcolorbox}[colback=white, boxrule=0.3mm]
    \centering 
      \footnotesize
    \begin{tabular}{p{0.97\columnwidth} c}
    \VarSty{ {\bf Tactile Sensation Descriptions} } & \hspace{-5cm} \multirow{5}{*}{ \includegraphics[height=1.5cm]{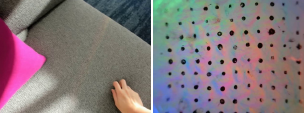} } \\
    \parbox[t]{0.68\columnwidth}{Sentence description: The object being touched is a synthetic fabric sofa, with the point of contact being the seat cushion. The cushion feels tightly woven like textile, displaying a uniform texture, and it is soft and resilient.} & \\
    \parbox[t]{0.6\columnwidth}{Phrase description: slightly rough texture, firm resistance, delicate comfort} & \\
    \hrulefill &  \\
    \VarSty{ {\bf Conversation } } & \\
   Question: How does the texture feel?  & \\
   Answer: The texture is slightly rough, but not unpleasant. It has a uniform woven pattern, giving it a consistent tactile sensation across the surface. & \\
   Question: Is it soft or hard? \\
   Answer: The surface feels soft and resilient. Pressing against it reveals that the underlying cushion offers firm resistance, but it quickly returns to its original shape, providing a delicate balance between comfort and support. & \\
   Question: How graspable is this object?
   Answer: The object is quite graspable. The slight roughness of the fabric provides a secure grip, and its soft yet firm structure makes it comfortable to hold or manipulate without slipping. & \\
   Question: How elastic is it? & \\
   Answer: The fabric itself has minimal elasticity, but the cushion underneath exhibits notable resilience. When pressed, the cushion compresses and then springs back smoothly, suggesting good elasticity in the underlying material. & \\
   Question: What is the most distinctive sensation you feel when touching this object? & \\
   Answer: The most distinctive sensation is the interplay of the slightly rough texture of the fabric with the soft but firm resistance of the cushion. This duality creates a feeling of robust comfort, making the object both tactilely engaging and supportive.
    \end{tabular}
\end{tcolorbox}
\vspace{-2mm}
\caption{One example from the example pool. The top block presents multi-granularity tactile sensation descriptions of the touched object, including both sentence-level and phrase-level descriptions, while the bottom block displays the corresponding group of conversations. Note that the visual and tactile images are not used as contexts and are only shown here for reference.
}
    \label{tab:example}
\end{minipage}
\end{table*}

The data construction process is as follows. First, building upon the properties and characteristics previously outlined, a set of exemplars is manually designed for various objects to be touched, thus constructing a pool of exemplars. As shown in Table~\ref{tab:example}, an example involves tactile sensation descriptions of a multi-granularity nature, as well as a group of conversations between a human and an assistant. In these conversations, the human asks questions to the assistant regarding the touched object, and the assistant responds as if they are physically interacting with the object. It is important to note that only questions with definitive answers are included. Since Touch100k provides comprehensive tactile sensation descriptions derived from visual images, the vision, touch, and language modalities within the dataset are semantically aligned. We use multi-granularity tactile sensation descriptions as input and dynamically select a subset from the exemplar pool as seed samples for in-context learning, which are then queried using GPT-4. The detailed prompt can be found in Table~\ref{tab:prompt}. Finally, we transform the query response into a single-turn conversation and conduct manual review and correction, resulting in a total of 5K unique touch-language instruction-following data.

\subsection{TactileBench Construction Details}
Since Touch100k \cite{cheng2024touch100k} does not contain a test set with tactile descriptions, we use the material classification test set from \textit{Touch and Go} \cite{yang2022touch} as our baseline data. This test set consists of visual images, tactile images, and classification labels. To construct tactile question-answering data for three types of tasks, we designed a unified prompt template using visual images and classification labels, queried GPT-4\footnote{All GPT-4 models mentioned in this paper are the gpt-4o-2024-11-20 version.}, and conducted manual verification. The prompt template is shown in Table~\ref{tab:benchmark_prompt}. 
Post-processing of query responses to generate single-turn conversation data is performed similarly to the construction of instruction data. Through extensive experimentation, we found that GPT-4 struggles to generate 3-5 long answers at once, instead breaking a single long answer into multiple parts. This behavior conflicts with the open-ended, free-form nature of our tactile commonsense reasoning conversations. To resolve this, we generate each of the 3-5 answers separately.

\begin{table*}[!htbp
]\centering
\begin{minipage}{0.99\linewidth}\vspace{0mm}    \centering
\begin{tcolorbox}[colback=white, boxrule=0.3mm]
    \centering
    \small
     \hspace{-6mm}
    \begin{tabular}{p{0.99\linewidth}}

\begin{minipage}{0.99\linewidth}\vspace{0mm}

\VarSty{messages} = [
            \{\var{"role":"system", "content":} f\var{"""}You are an AI assistant, and you are touching a single object. What you touch are provided with two description, describing the same object you are interacting with. Answer all questions as you are touching the object. \\

Design a conversation between you and a person asking about this object. The answers should be in a tone that a AI assistant is touching the object and answering the question. Ask diverse questions and give corresponding answers.\\

Include questions asking about the typical physical properties of the object being touched, including the material, texture, roughness, hardness, bumpiness, elasticity, weight, sharpness, extensibility, etc. Only include questions that have definite answers and can be answered confidently.\\

Also include complex questions that are relevant to the touched object, for example, asking about background knowledge of the touched object, or asking to discuss characteristics related to interaction and perception during touching, including graspability, tingling sensation, and the object’s flexibility, etc. Again, do not ask about uncertain details. Provide detailed answers when answering complex questions. For example, give detailed examples or reasoning steps to make the content more convincing and well-organized. You can include multiple paragraphs if necessary.\var{"""}\}] \\
    
    \For{ \var{\VarSty{example} \textnormal{in} \VarSty{fewshot\_examples}}}
    {
         \var{\VarSty{messages}.append(\{"role":"user", "content":\VarSty{ example[`context']}\})} \; \\
         \var{\VarSty{messages}.append(\{"role":"assistant", "content":\VarSty{example[`response']}\} ) } \;
         }  
    \var{\VarSty{messages}.append(\{"role":"user", "content":`\textbackslash  n'.join(\VarSty{query})\})}

\end{minipage}
    \end{tabular}
\end{tcolorbox}
    
\vspace{-2mm}
\caption{For each query, we demonstrate the prompt construction process for GPT-4 
 to collect \VarSty{ query[`response']}  from \VarSty{ query[`context']} using few-shot in-context learning, where \VarSty{fewshot\_examples} are dynamically select from the example pool, with each example consisting of input \VarSty{example[`context']} and output \VarSty{example[`response']}. }
    \label{tab:prompt}
\end{minipage}
\end{table*}

\begin{table*}[!htbp]\centering
\begin{minipage}{1.0\linewidth}\vspace{0mm}    \centering
\begin{tcolorbox}[colback=white, boxrule=0.3mm]
    \centering 
      \footnotesize
    \begin{tabular}{p{0.97\columnwidth} c}
    
    \VarSty{ {\bf Context Type 1: Text Prompt } } &  \\ 
    \parbox[t]{\columnwidth}{You are an AI visual assistant looking at a single image, where you can see the object being touched is \texttt{<Object Class>}. Imagine you are physically touching the object being touched in the image. \newline
    \newline Design conversations between you and a person asking about this object. The answers should be in a tone that you is touching the object and answering the question. \newline \newline
    \texttt{<Task Specific Prompt>}
    } & \\
    \\

    \VarSty{ {\bf Context Type 2: Visual Image} } &  \\ 
    \parbox[t]{\columnwidth}{ \multirow{5}{*}{ \includegraphics[height=1.5cm]{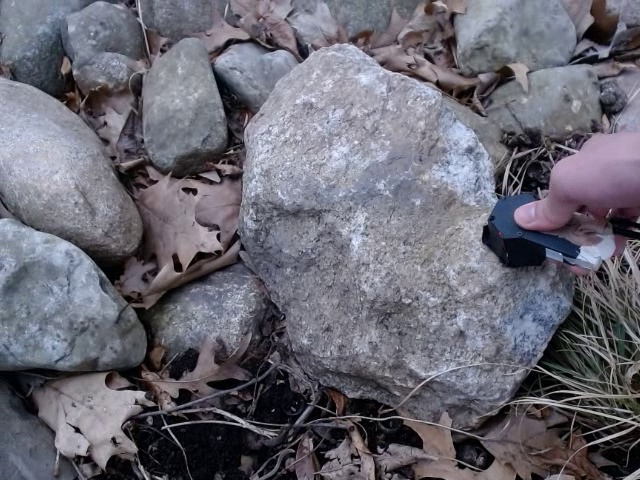} } 
    \newline
    \newline
    \newline
    \newline
    }& \\
    
    \hrulefill &  \\
    \VarSty{ {\bf Object Class } } & \\
    rock & \\
    \VarSty{ {\bf Task Specific Prompt Type 1: Fundamental Attributes Understanding }  } & \\
   Include questions asking about the typical physical properties of the object being touched, including but not limited to the material, texture, roughness, hardness, bumpiness, elasticity, weight, sharpness, extensibility. Randomly select five of these physical properties, and design a separate conversation for each. Ensure that the conversation contains only the confirmed information. & \\ 
   Required format: & \\
\textnormal{[}\{``question": 
\texttt{<question 1>} , ``answer": \texttt{<answer>}\}, & \\
\{``question": 
\texttt{<question 2>} , ``answer": \texttt{<answer>}\}, & \\
\{``question": 
\texttt{<question 3>} , ``answer": \texttt{<answer>}\}, & \\
\{``question": 
\texttt{<question 4>} , ``answer": \texttt{<answer>}\}, & \\
\{``question": 
\texttt{<question 5>} , ``answer": \texttt{<answer>}\}\textnormal{]} & \\

\VarSty{{\bf Task Specific Prompt Type 2: Tactile Interaction Perception } } & \\
   Include questions asking about the characteristics related to interaction and perception during touching, including but not limited to the graspability, prickliness, bendability, and malleability. Randomly select three of these interaction and perception characteristics, and design a separate conversation for each. Each conversation should inclue one question and 3-5 corresponding answers as ground truth. Ensure that the conversation contains only the confirmed information. & \\ 
   Required format: & \\
\textnormal{[}\{``question": 
\texttt{<question 1>} , ``answer": \texttt{<answer>}\}, & \\
\{``question": 
\texttt{<question 2>} , ``answer": \texttt{<answer>}\}, & \\
\{``question": 
\texttt{<question 3>} , ``answer": \texttt{<answer>}\}\textnormal{]} & \\
\VarSty{ {\bf Task Specific Prompt Type 3: Background Knowledge Reasoning} } & \\
   Include complex reasoning questions involving background knowledge related to the touched object, emphasizing the tactile perspective.Design two conversations, Each with one question and 3-5 candidate answers as ground truth, with each answer being independent and self-contained. Ensure that the conversation contains only the confirmed information. Provide detailed answers when answering complex questions. For example, give detailed examples or reasoning steps to make the content more convincing and well-organized. You can include multiple paragraphs if necessary. & \\ 
   Required format: & \\
\textnormal{[}\{``question": 
\texttt{<question 1>} , ``answer": \texttt{<answer>}\}, & \\
\{``question": 
\texttt{<question 2>} , ``answer": \texttt{<answer>}\}\textnormal{]} & \\
   
    \end{tabular}
\end{tcolorbox}
\vspace{-2mm}
\caption{Prompt template for building our benchmark. The top block presents the context fed into GPT-4, which consists of both a text prompt and a visual image. Notably, the \texttt{<object class>} and \texttt{<Task specific Prompt>} in the text prompt are detailed in the bottom block. It is worth mentioning that the responses obtained from GPT-4 undergoes further processing.
}
    \label{tab:benchmark_prompt}
\end{minipage}
\end{table*}

\begin{table}[]
\centering
\begin{small}
\begin{tabular}{lcc}
\toprule
\textbf{Configuration} & \textbf{Stage I} & \textbf{Stage II}  \\
\midrule
Optimizer & Adam & Adam  \\
Learning Rate & 5e-4 & 2e-5  \\
Weight Decay & 0.001 & 0.001  \\
Training Epochs & 1 & 1  \\
Warmup Ratio &  0.1 &  0.1  \\
Learning Rate Scheduler &  Linear &  Linear  \\
Batch Size Per GPU & 16 & 16  \\
Maximum Token Length & 512 & 512  \\
Unfreeze LLM & \ding{55} & \ding{51} \\
Enable MoE & \ding{55} & \ding{51} \\
\bottomrule
\end{tabular}
\end{small}
\caption{\textbf{Training configuration of \model.} 
Stage I: Tactile Token Adaption for LLM. Stage II: End-to-end Fine-tuning with MoE.}
\label{tab:param_config}
\end{table}

\subsection{Training Configuration}
We present the detailed training configuration and hyper-parameters in the two stages in Table~\ref{tab:param_config}.

\subsection{GPT-4 and DeepSeek-R1 scoring}
Since TactileBench is an open-ended, free-form tactile commonsense reasoning dataset, traditional n-gram-based metrics such as CIDEr \cite{vedantam2015cider} and B@4 \cite{papineni2002bleu} are not suitable. Instead, we evaluate the dataset using the METEOR \cite{banerjee2005meteor} metric, which is based on semantic similarity, along with multi-dimensional scoring from GPT-4 \cite{achiam2023gpt} and DeepSeek–R1 \cite{guo2025deepseek}. 

Both GPT-4 and DeepSeek–R1 raters evaluate responses across four key dimensions: accuracy, relevance, completeness, and fluency. Specifically, raters are prompted to report a 1-10 score and a corresponding rationale for each of these four dimensions. The final score is computed as the average of the four individual dimension scores. The evaluation criteria for each of the four dimensions are summarized as follows:
\begin{itemize}
\setlength{\itemsep}{0pt}
\setlength{\parsep}{0pt}
\setlength{\parskip}{0pt}
  \item \textit{Accuracy:}   Does the answer match the reference answer? Are there any factual errors?
  \item \textit{Relevance:} Does the answer directly address the question, without redundancy or deviation from the content?
  \item \textit{Completeness:} Does the answer cover all key information? Are there any important details missing? 
  \item \textit{Fluency:} Is the language natural and smooth, with no grammatical errors? 
\end{itemize}

\end{document}